\documentclass[conference]{IEEEtran}
\IEEEoverridecommandlockouts
\usepackage{cite}
\usepackage{amsmath,amssymb,amsfonts}
\usepackage{graphicx}
\usepackage{textcomp}
\usepackage{xcolor}
\usepackage{hyperref}
\usepackage{fancyhdr}
\usepackage{algorithm} 
\usepackage{algpseudocode}

\algnewcommand\algorithmicinput{\textbf{Input:}}
\algnewcommand\Input{\item[\algorithmicinput]}

\algnewcommand\algorithmicoutput{\textbf{Output:}}
\algnewcommand\Output{\item[\algorithmicoutput]}
\usepackage{pifont}
\newcommand{\cmark}{\textcolor{blue}{\ding{51}}}%
\newcommand{\xmark}{\textcolor{red}{\ding{55}}}%
\def\BibTeX{{\rm B\kern-.05em{\sc i\kern-.025em b}\kern-.08em
    T\kern-.1667em\lower.7ex\hbox{E}\kern-.125emX}}

\title{\huge SILOP: An Automated Framework for Semantic
Segmentation Using Image Labels Based on Object Perimeters\\
}

\author{
    \IEEEauthorblockN{
    Erik Ostrowski\IEEEauthorrefmark{1}, Bharath Srinivas Prabakaran\IEEEauthorrefmark{1}, Muhammad Shafique\IEEEauthorrefmark{3}}
    \IEEEauthorblockA{\IEEEauthorrefmark{1}Institute of Computer Engineering, Technische Universit{\"a}t Wien (TU Wien), Austria
    \\ \{erik.ostrowski, bharath.prabakaran\}@tuwien.ac.at}
    \IEEEauthorblockA{\IEEEauthorrefmark{3}Division of Engineering, New York University Abu Dhabi (NYUAD), United Arab Emirates (UAE)
    \\muhammad.shafique@nyu.edu}
}

\fancypagestyle{firstpage}
{
    \fancyhead[C]{To appear at the International Joint Conference on Neural Networks (IJCNN), July 2023, Gold Coast, Queensland, Australia.}    
    \fancyhead[R]{}
}

\begin{document}


\maketitle

\thispagestyle{firstpage}

\begin{abstract}
Achieving high-quality semantic segmentation predictions using only image-level labels enables a new level of real-world applicability.
Although state-of-the-art networks deliver reliable predictions, the amount of handcrafted pixel-wise annotations to enable these results are not feasible in many real-world applications.
Hence, several works have already targeted this bottleneck, using classifier-based networks like Class Activation Maps~\cite{CAM} (CAMs) as a base.
Addressing CAM's weaknesses of fuzzy borders and incomplete predictions, state-of-the-art approaches rely only on adding regulations to the classifier loss or using pixel-similarity-based refinement after the fact.
We propose a framework that introduces an additional module using object perimeters for improved saliency.
We define object perimeter information as the line separating the object and background.
Our new PerimeterFit module will be applied to pre-refine the CAM predictions before using the pixel-similarity-based network.
In this way, our PerimeterFit increases the quality of the CAM prediction while simultaneously improving the false negative rate.
We investigated a wide range of state-of-the-art unsupervised semantic segmentation networks and edge detection techniques to create useful perimeter maps, which enable our framework to predict object locations with sharper perimeters.
We achieved up to 1.5\% improvement over frameworks without our PerimeterFit module.
We conduct an exhaustive analysis to illustrate that SILOP enhances existing state-of-the-art frameworks for image-level-based semantic segmentation.
The framework is open-source and accessible online at \url{https://github.com/ErikOstrowski/SILOP}.
\end{abstract}



\section{Introduction}

\begin{figure*}[ht]
\centering
\includegraphics[width=150mm,scale=0.50]{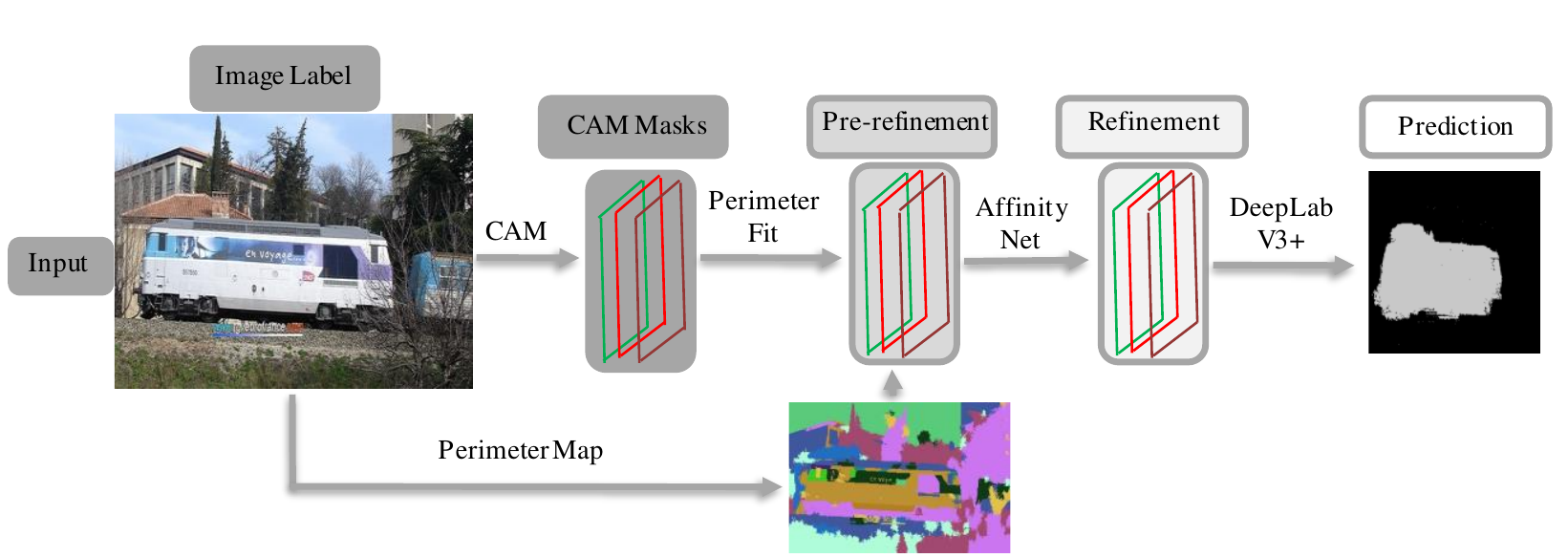} 
\caption{A high-level overview of the SILOP framework training, starting with the input and a CAM to create the first predictions, then our PerimeterFit module pre-refines the input with the help of the perimeter map, then AffinityNet performs the final refinement, and a fully supervised network uses the refined masks as labels.\label{flow}}
\end{figure*}



Semantic segmentation describes the task of assigning every pixel in a given image to a class.
In contrast to assigning the whole image to a class, pixel-wise classification offers much more information about the image's content and can be used for many more applications.
Those applications range from autonomous and self-driving vehicles~\cite{AD}, remote sensing~\cite{diakogiannis2020resunet, kemker2018algorithms}, facial recognition~\cite{meenpal2019facial, khan2015multi}, agriculture~\cite{milioto2018real, barth2018data}, and in the medical field \cite{rehman2020deep, zhao2017tracking}, etc.

The use of deep learning for semantic segmentation greatly advanced the quality of the state-of-the-art and achieved new top scores at incredible speed since then.
Nevertheless, state-of-the-art in semantic segmentation is exclusively 
achieved through fully supervised networks.
Fully supervised means that during training, the networks have access to ground truth masks where all pixels are optimally classified.
However, the problem with this approach is that the optimal ground truths are created by hand, and we usually need over $10,000$ annotations to achieve state-of-the-art predictions.
Therefore, creating such datasets is tedious and time-consuming at best when dealing with a scenario where basically anyone can do this work \cite{CT}.
But as soon as we move to more sophisticated applications, like brain tumor detection, creating the annotations by hand becomes much more complex, and experts capable of doing it become much more expensive and less available.
Additionally, we experience annotator bias in some fields, where even experts argue about where to draw the correct line between the target object and background.
Hence, this paper will explore the possibility of performing semantic segmentation with just image-level annotation.

\begin{figure}[ht]
\centering
\includegraphics[width=0.9\columnwidth]{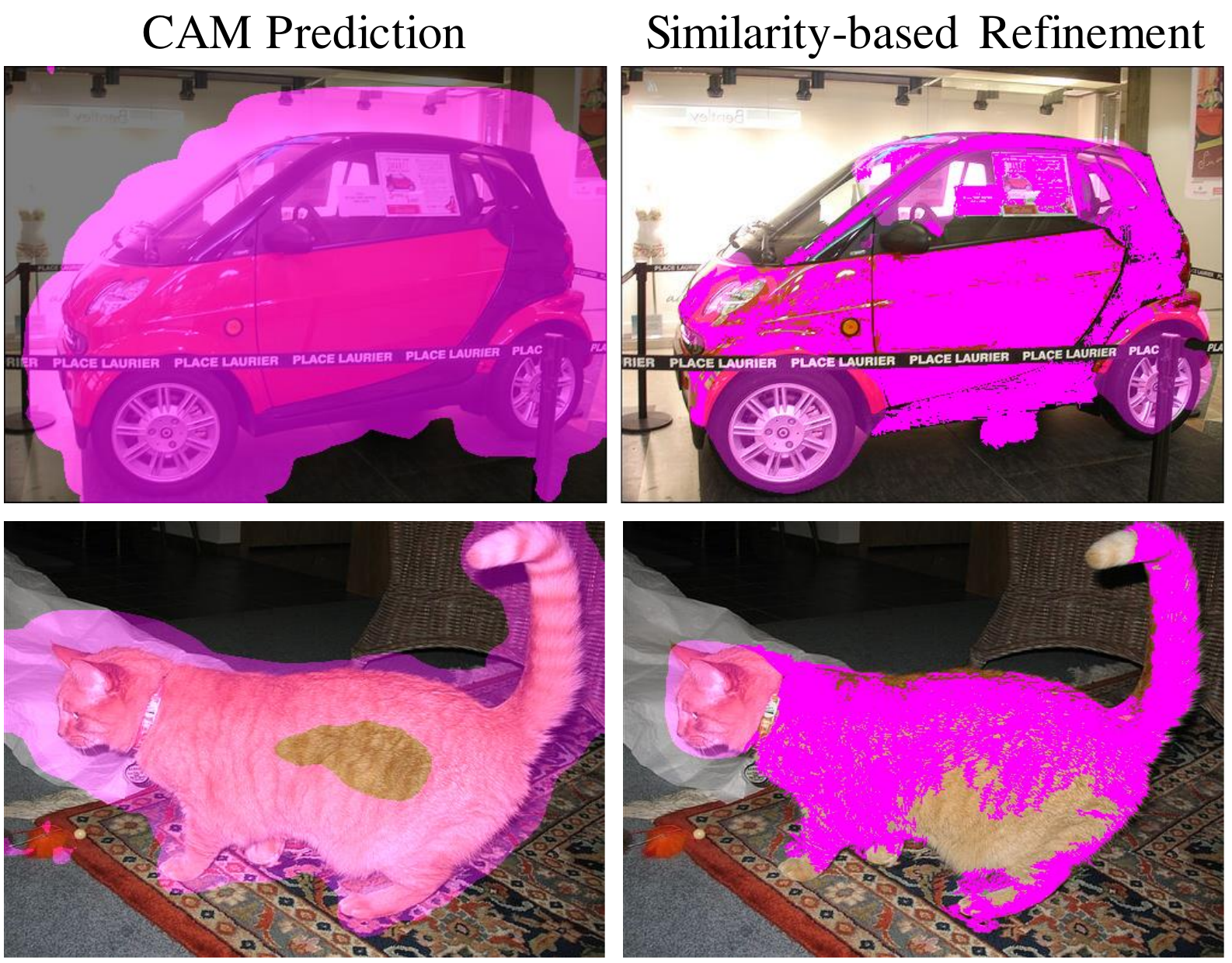} 
\caption{Comparing the masks from CAM prediction and pixel-similarity based refinement.\label{CAMs}}
\end{figure}

Most state-of-the-art approaches to semantic segmentation using only image-level labels are based on Class Activation Maps (CAMs)~\cite{CAM}.
CAMs are constructed with convolutional neural networks as a backbone.
The first CAMs added to the backbone a final convolutional layer with as many channels as classes, and a Global Average Pooling (GAP) module returned then a classification score.
This way, the modified network could be trained like a conventional classification model.
But if we take the output before the GAP, we have feature maps that show which part of the input image was mostly responsible for the classification decision and hence can be used as a pixel-wise prediction, fig.~\ref{fig2} section A presents an overview of this process.
The downside of this approach is that normal classifiers cannot give a detailed pixel-wise prediction as a fully supervised semantic segmentation network could.
CAMs tend to highlight the most descriptive part of an image, for example, the body of a car, while ignoring the rest of the target object, for example, the windows or tires.
Therefore, research focuses on using the CAM predictions as a base for improving it after the fact or changing the loss function to guide the classifier to predict more complete masks.
Works that focus more on refining the initial mask often use pixel similarity-based methods that reclassify pixels based on the similarities and classes of the neighboring pixels.

Whereas works that try to improve the original CAM output focus on adding regularizations to the classification loss, like punishing the network if CAMs of parts of the image do not fit CAMs of the whole image.
Nevertheless, both approaches have their weaknesses.
The regulation-based methods, for example, may succeed in predicting the complete target object, but they cannot detect sharp borders like the fully supervised networks.
On the other hand, pixel-similarity-based methods can generate sharp borders.
Still, they cannot extend the prediction to the complete target object if the base mask does not cover it and the other parts of the object have a different color, like the car body, wheels, and windows.
Hence, state-of-the-art approaches combine both methods.
However, we noticed that none of the state-of-the-art methods utilize additional information like edge detection that can give the CAM crucial information on how to shape the initial prediction.
Therefore, we propose our framework that introduces a pre-refinement step between CAM and pixel-similarity-based refinement.
In our pre-refinement step, we will combine conventional edge detection and image clustering to create a perimeter map, which can adapt the CAM prediction to the borders of a target object.
Fig.~\ref{flow} presents a high-level overview of SILOP.
Our experiments have shown that a higher-quality CAM greatly benefits the subsequent use of pixel-similarity-based refinement methods.
We perform extensive experiments on the PASCAL VOC2012 dataset~\cite{VOC} to prove the effectiveness of the proposed framework in various experimental settings and compare them with a wide range of state-of-the-art techniques to illustrate the benefits of our approach.

\textbf{The key contributions} of this work are:
\begin{enumerate}
    \item Our SILOP framework improves the prediction quality of the segmentation mask by introducing an additional step between CAM and pixel similarity-based refinement.
   \item Our PerimeterFit module can be incorporated into any conventional CAM framework and already working WSSS pipelines. Since the PerimeterFit module is only used for the generation of labels, it won't add more computations for inference predictions.
   \item We have presented detailed ablation studies and analysis of the results compared our framework to state-of-the-art methods on the VOC2012 dataset to evaluate our method's efficacy and the improvements achieved using our framework.
   \item The complete framework, including the novel PerimeterFit module, is open-source and accessible online at \url{https://github.com/ErikOstrowski/SILOP}.
\end{enumerate}

\section{Related Works}

This section discusses the current state-of-the-art class activation maps and pixel similarity methods.
 





\subsection{Classical CAMs}

The class activation map (CAM) method is a fundamental approach to generating semantic segmentation masks from image-level labels.
The central idea of CAMs is to use any model trained with classification loss to generate activation maps that highlight the image regions responsible for the prediction decision.
This results in a rough localization of the objects rather than precise pixel-wise masks.

Several approaches extend on CAM to improve its prediction quality.
The most popular CAM approaches focus on adding regularization loss to improve the quality of the CAM prediction~\cite{SUB, PUZZLE} or utilizing refinement methods that aim to enhance the CAM afterward~\cite{AFF, SEAM}.
For example, PuzzleCAM~\cite{PUZZLE} introduces a regularization loss by sub-dividing the input image into multiple parts, forcing the network to predict image segments that contain the non-discriminative parts of an object.
AdvCAM~\cite{AdvCAM} reverses adversarial attacks to guide the classifier to detect the complete object.
Additionally, Lee et al. introduced a regularization to punish activations outside the adversarial highlighted areas.
CLIMS~\cite{CLIMS} trained the network by matching text labels to the correct image.
Hence, the network tries to maximize and minimize the distance between correct and wrong pairs, respectively, instead of just giving a binary classification result.

\subsection{CAMs and pixel-similarity techniques}

Since the methods focusing on improving CAM loss still generate blurry masks, other approaches aim to refine the CAM afterward.
SEAM~\cite{SEAM} proposed a combination of regularization and pixel similarity to create higher-quality predictions.
The regularization loss compares a response map generated by the original image to a response map generated by an affine-transformed image to improve the model's generalization ability.
Additionally, the similarity module searches for pixels similar to already positively classified pixels to enlarge the first prediction.
AffinityNet~\cite{AFF} trains a second network to learn pixel similarities, which generates a transition matrix combined with the CAM iteratively to refine its activation coverage.

\begin{table}[ht]
\caption{Comparison of state-of-the-art semantic segmentation techniques using image labels.\label{check}}

\centering 
{
\begin{tabular}{lcccc}
\hline
Methods& \begin{tabular}{@{}c@{}}Loss/\\ Regularization\end{tabular}     & \begin{tabular}{@{}c@{}}Pixel\\ Similarity \end{tabular} & \begin{tabular}{@{}c@{}}Object\\ Perimeters\end{tabular} \\ \hline
Jiang et al.~\cite{OOA}              & \cmark  &   \xmark &   \xmark \\
AffinityNet~\cite{AFF},  & \xmark  &   \cmark &   \xmark  \\
Ahn et al.~\cite{IRNET}  & \xmark  &   \cmark &   \xmark  \\
Fan et al.~\cite{ICD}  & \xmark  &   \cmark &   \xmark  \\
PuzzleCAM~\cite{PUZZLE} & \cmark  &   \cmark &   \xmark  \\
Xie et al.~\cite{CLIMS} & \cmark  &   \cmark &   \xmark  \\
Lee et al.~\cite{AdvCAM} & \cmark  &   \cmark &   \xmark  \\
Wang et al.~\cite{SEAM} & \cmark  &   \cmark &   \xmark  \\
Lee et al.~\cite{FICKLE} & \cmark  &   \cmark &   \xmark  \\
Chang et al.~\cite{SUB} & \cmark  &   \cmark &   \xmark  \\
Wang et al.\cite{ MCOF} & \xmark  &   \cmark &   \cmark  \\\hline
SILOP (Ours)              & \cmark  &   \cmark &   \cmark  \\
\end{tabular}}
\end{table}


Table.~\ref{check} highlights the major sources of improvement of state-of-the-art methods.
We observe only one method just used additional regularization for the loss function, whereas several focused on improving the CAM afterwards.
However, most state-of-the-art approaches rely on a combination of AffinityNet for adding pixel-similarity-based refinement, and an improved loss function.
Moreover, we note that Wang et al.~\cite{ MCOF} did utilize a kind of perimeter-based approach, but without many of the additional steps SILOP uses to maximize the utility.

\section{Our Framework}

\begin{figure*}[ht]
\centering
\includegraphics[width=150mm,scale=0.50]{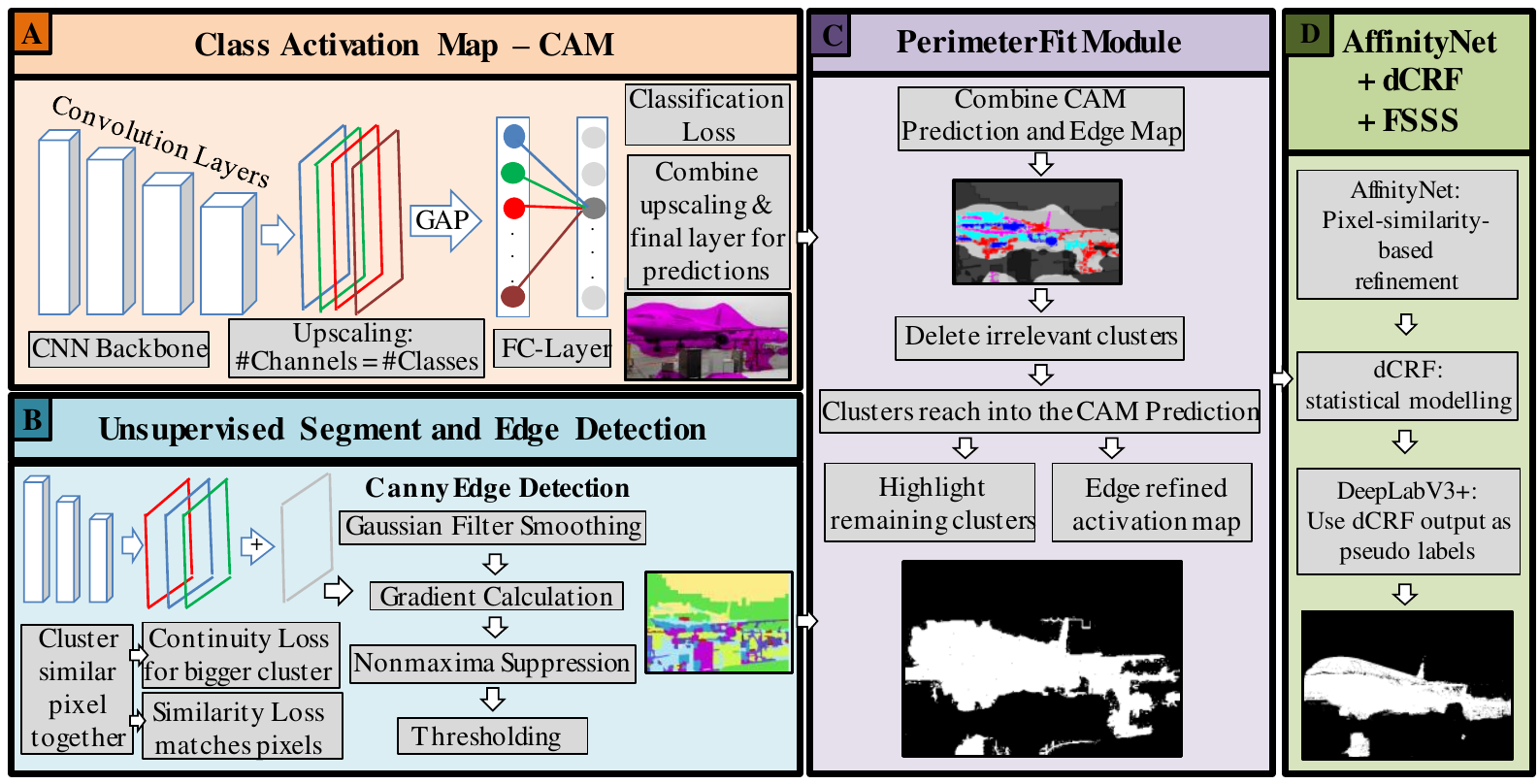} 
\caption{Overview of the SILOP framework. 
(A) Stages in the class activation map - CAM, which is used to generate an initial response map;
(B) The USS stages cluster the pixel images into bigger groups to generate a perimeter map for the object; 
(C) Our PerimeterFit module conforms the initial response map to the generated map to obtain fine-grained masks for each object; 
(D) AffinityNet uses the masks to generate pseudo-labels, which can be used to train a DNN model for FSSS. \label{fig2}}
\end{figure*}

Fig.~\ref{fig2} presents an overview of the SILOP framework.
We start with step A, creating rough masks with a CAM.
We applied the PuzzleCAM method to create the base masks for our framework.
 In step B, we simplify the input images by clustering pixels together.
Again, any state-of-the-art clustering method will work in our framework, but we applied variations of an unsupervised segmentation (USS) network.
The USS is followed up by a deterministic edge detection algorithm that outlines the object's perimeter in the simplified image to generate a perimeter map for the input image.
In step C, we apply our PerimeterFit module that removes all positive classifications outside of our perimeter map.
Finally, the AffinityNet~\cite{AFF} uses the output masks from the PerimeterFit module to generate the pseudo-labels, which are further refined by additional minor methods like dense Conditional Random Fields (dCRF)~\cite{CRF}.
In the last step, we use the pseudo-labels to train a fully supervised DeepLabV3+ network~\cite{DEEP} to analyze our framework's benefits and illustrate its efficacy.

\subsection{Class Activation Map - CAM}

SILOP starts with the generation of an initial response map using a CAM model, which given an input image $I$, will return $n$ masks.
We will follow the established approach of retrieving only masks of classes present in $I$ rather than relying on correct classifications.
Nevertheless, the CAM predictions have fuzzy borders and, for the most part, highlight only parts of the object, as illustrated in Fig.~\ref{CAMs}.

For example, the initial response map for a cat would cover the head, parts of the body, and a significant amount of background pixels, but it does not include the legs or the tail.
To address this issue, we will use PuzzleCAM, which uses a modified loss function that motivates the model to focus more on other parts of the object during training.
Note that the use of other CAM approaches is orthogonal to PuzzleCAM. 
Any such state-of-the-art solution, with similar interfaces to PuzzleCAM, can be easily replaced in our framework due to its dexterity in achieving the required results.

\subsection{USS and Edge Detection}

\begin{figure}[t]
\centering
\includegraphics[width=85mm,scale=0.85]{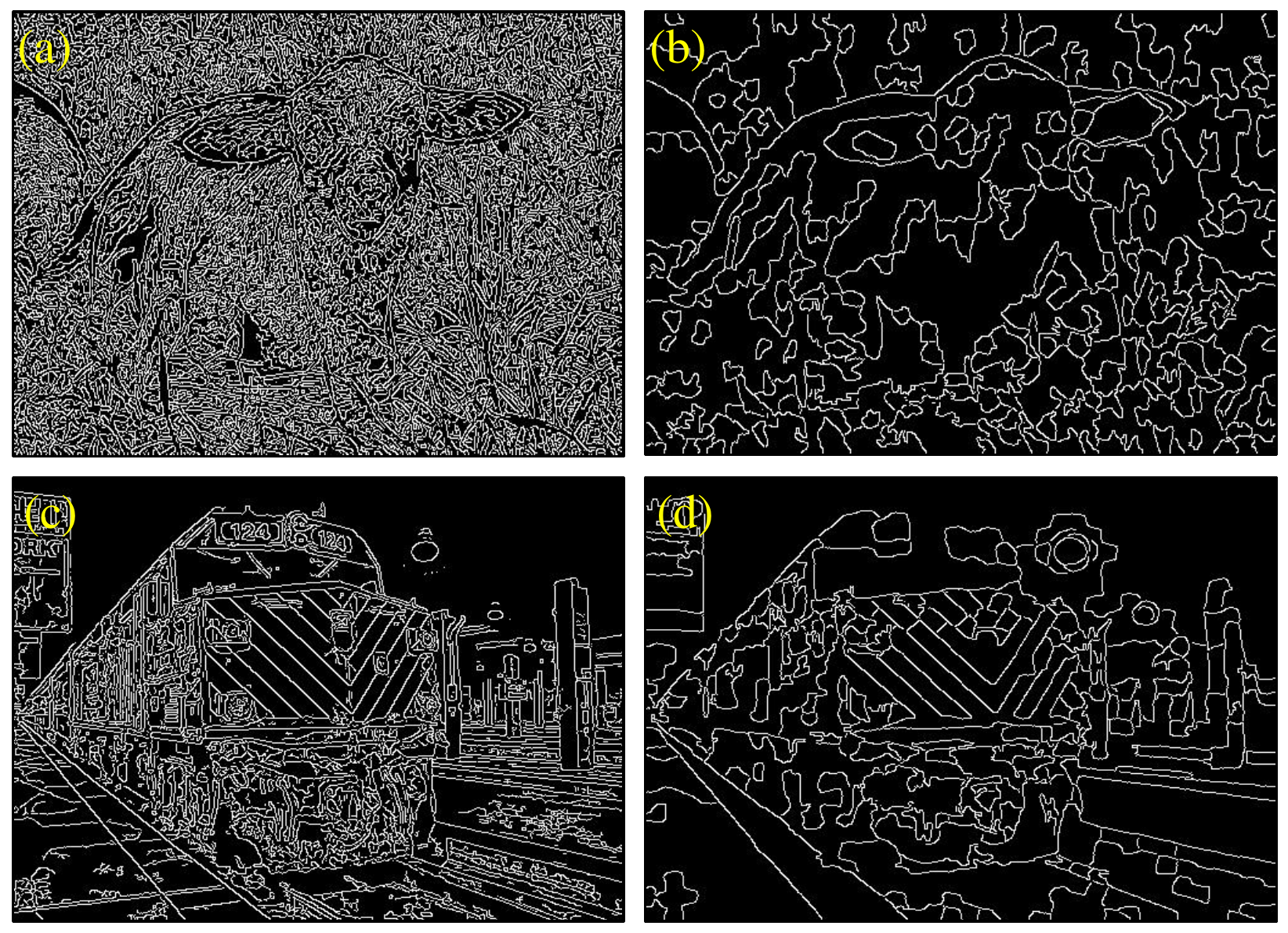} 
\caption{Analyzing the effect of USS on images. (a) and (c) illustrate the use of edge detection on the original images, whereas (b) and (d) illustrate the decrease in edges when USS-generated perimeter maps are used as input.\label{edge}}
\end{figure}

Conventionally, the next step would already include the usage of a pixel-similarity-based refinement method, mostly AffinityNet, to extract a fine-grained segmentation mask.
Instead, we propose to take an intermediate step to improve the CAM's quality.
Higher-quality CAMs will transfer to higher-quality AffinityNet output since AffinityNet needs high-certainty anchors from which it extends the mask.
Nevertheless, just using overall higher-quality CAM may not improve AffinityNet but the CAMs need to be balanced between high overall quality and a low False Positive (FP) rate.
A low FP rate means that the prediction does not include unrelated image parts to the target object, a drawback of the conventional CAM approaches, as discussed earlier.
Therefore, we aim to create a perimeter map that can be used to eliminate unrelated activations to the target object.
Unfortunately, applying an edge detection method to the input image is insufficient to generate the necessary perimeter map required for this approach.
Without any pre-processing, the perimeter map can be too fine-grained, which leads to minimal changes in the final mask when reducing the masks to the nearest perimeters.
On the other hand, coarse-grained maps can result in a highly sparse perimeter map, where a reduction to the detected perimeters could lead to the truncation of significant object parts or the removal of the mask altogether. Fig.~\ref{edge} illustrates some of these problems we encountered when using edge detection without USS.

Hence, we aim to simplify the original image so that the edge detector consistently outputs more useful perimeter maps.
For this purpose, we chose a clustering method that groups image pixels together meaningfully, so we do not lose the target object's shape.
We used Kanezaki et al.'s ~\cite{USS} method for our experiments, consisting of two convolutional layers, ReLU activation and Batch Norm.
Furthermore, \cite{USS} uses two loss functions.
We performed exhaustive experiments to find a set of hyperparameters for the first loss so that, on average, the degree of simplification results in beneficial perimeter map output.
We combined the results of using SLIC~\cite{SLIC} and Quickshift~\cite{QUICK} for the first loss.
The second loss limits the number of clusters generated to a maximum value of $q$.
A low $q$ value results in a very coarse simplification with next to no objects left, while a high $q$ value results in little simplification of the original image.
The batch normalization evens the probability for each $\hat{q}$ out.

Subsequently, applying edge detection to the simplified image offers the best balance between keeping the essential object edges while not including too many nonessential edges.
We use the Canny Edge Detection technique~\cite{CANNY}, which consists of Gaussian Filter Smoothing, Gradient calculation, Non-Maxima Suppression, and Thresholding.
In our experiments, the Canny method offered the best balance between keeping the essential object edges while not including too many nonessential edges.

\begin{algorithm}[t]
\caption{Step B}\label{alg:stepB}
\begin{algorithmic}[1]
\Input{ImageList, q, Quickshift, SLIC}
\Output{PerimeterMapList}
\For {$I$ \textbf{in} ImageList}
\If {Quickshift}
\State $I_{USS}$ $\leftarrow$ $USS_{Quickshift}(I, q)$
\EndIf
\If{SLIC}
\State $I_{USS}$ $\leftarrow$ $USS_{SLIC}(I, q)$
\EndIf
\State USSList.append($I_{USS}$)
\EndFor
  
\For {$I_{USS}$ \textbf{in} USSList}
\State $I_{PM}$ $\leftarrow$ CannyEdgeDetection($I_{USS}$)
\State PerimeterMapList.append($I_{PM}$)
\EndFor
\end{algorithmic}
\end{algorithm}

\subsection{PerimeterFit Module}

\begin{algorithm}[t]
\caption{PerimeterFit}\label{alg:PF}
\begin{algorithmic}[1]
\Input{PerimeterMapList, CAMList, Threshold}
\Output{PredictionList}
		\For { $CAM,PM$ \textbf{in} CAMList, PerimeterMapList}
            \For { $(x,y)$ \textbf{in} CAM}
                \If {$I(x,y) \leq$ Threshold}
                        \State NB $\leftarrow$ neighbors of$(x,y)$
                        \While {not($\forall$ (PM(NB)=$255$ or PM(NB)=$0$))}
                        
                            \State PM(neighbors) $\leftarrow 0$
                            \State $(x,y)$ $\leftarrow$ neighbors
                        \EndWhile
                \EndIf  
            \EndFor
            \State PredictionList.append(Prediction)
		\EndFor
\end{algorithmic}
\end{algorithm}

In step C, we will apply our PerimeterFit module in combination with the perimeter map to refine the rough CAMs from step A.
The refinement process determines which positively classified pixels will be kept while removing lower certainty positives.
Note that with this approach, we can only shrink the original CAM size and not extend it as AffinityNet would.
We will utilize the perimeter map from step B to decide which positive classified pixels in the CAM are more likely to be correct.
We achieve this by utilizing the rough CAM from step A to differentiate the object's perimeter from the remaining edges.
All positive classified pixels inside the assumed perimeter of the target object should remain in the mask, and the rest should be reclassified as background.
The reclassification will be done by a Floodfill algorithm~\cite{FLOOD}.
In detail, let us define a target cluster as the pixels in the image encircled by either edges in the Perimeter Map and the image border or only the edges.
Then, if any target cluster contains at least one pixel that is classified as background, the Floodfill will reclassify all pixels inside the target cluster as background.
After the Floodfill terminates, the only target clusters still classified as positive are inside the assumed perimeter of our target object.
Afterward, the refined masks will serve as pseudo-labels of AffinityNet.
As already mentioned, when describing step B, in our final version of the framework, we will use refined masks produced using SLIC and Quickshift.
Our experiments showed that we generated the best results when combining them, using a different threshold for each.
Hence, thresholds are essential hyperparameters for our framework because they are a major factor in balancing the overall quality of the CAM and its FP rate.
We estimated the most optimal thresholds through a grid search and evaluations done on the training set.

\subsection{AffinityNet and FSSS network}

After we fit the initial CAM prediction to our estimated perimeter map, we are ready to train the AffinityNet.
AffinityNet is composed of a random walk method that utilizes pixel affinities.
Starting from a set of positively and negatively classified pixels in the response map, the random walk uses its affinity score to reclassify pixels.
This methodology works great in cases where the initial response map contains only one part of the object while the rest differs little from the recognized part.

However, the random walk strategy fails to include dissimilar object parts, such as the wheels, not included in the original response map, as shown in Fig.~\ref{CAMs}.
On the other hand, given a bad initial response map, AffinityNet may further extend the false positives by including irrelevant background pixels as part of the object of interest.
To circumvent this scenario, we use only masks with a low false-positive rate for AffinityNet.
Furthermore, as common practice, we use dCRF to the response maps of AffinityNet to better capture the object's perimeters.

The dense Conditional Random Field is a discriminative statistical modeling method that models the prediction mask in a graph where each pixel is a node with the label information.
A CRF is dense if all nodes are connected.
After the graph is created, the algorithm changes the node labels to minimize an energy function that can be considered a loss function in this context.
Finally, we use the refined response maps as pseudo-ground truths to train a fully supervised semantic segmentation (FSSS) network.
We use the DeepLab-V3+ model with the ResNeSt-101 architecture~\cite{RESNEST} as the backbone model, like PuzzleCAM~\cite{PUZZLE}.
Since we have deliberately ignored classifying the objects in previous steps to reduce the risk of missing an object's mask, it is necessary to train an FSSS model with the predictions of the last stage as pseudo ground truths to be able to infer images without class labels.
This approach has been shown to generate higher-quality results than simultaneously having the CAM do the correct classification and segmentation.
Fig.~\ref{fig2} illustrates an overview of the process.

Let us assume that the initial prediction ${M}$ will not cover the whole object while misclassifying a certain amount of background pixels since a perfect segmentation is very unlikely.
Then we can divide the initial prediction into a sum of the set of positive pixels $P$ in the ground truth multiplied by an error rate $\epsilon \in [0,1]$ and the set of false-positive pixel predictions $FP$ as:
{
\begin{equation}
F_{CAM}(I) = M = P \epsilon + FP
\end{equation}}
Generally, based on experimental results, we have observed that negatively predicted pixels are determined with high confidence, whereas the confidence of positive pixel predictions is comparatively low.
Therefore, our method aims to reduce the number of FP pixels at the cost of increasing the FN rate, which increases the likelihood that a positive pixel is a TP.
Since AffinityNet uses a set of positive pixels and negative pixels as anchors to determine if the pixels in between are more like the positive or negative pixel set, it is of utmost importance to ensure that the anchors are correctly classified, i.e., true-positive, or true-negative.

On the one hand, we observed that this method excels when using images with very few objects of different colors, e.g., a green plane in the blue sky.
Then the USS will predict very few clusters and not merge objects with the background.
By using a low threshold prediction, ensuring that we cover the whole plane, PerimeterFit removes the sky cluster resulting in a fine-grained plane mask.
On the other hand, our approach struggles with images composed of many similarly colored and overlapping objects because the overlapping objects will most likely be covered entirely by the initial CAM.
In this case, those additional objects might still be part of the refined mask.
More significant problematic scenarios arise when the perimeter map cannot determine the object's perimeter correctly because of its color similarity to the background.
In such a case, the refinement process will sometimes delete a substantial part of correctly classified objects.
We also tested a second approach, using the Floodfill algorithm, wherein the objective was not to remove background pixels but to add foreground pixels.
Unfortunately, this approach does not improve the prediction accuracy since we need to expand from our original goal of identifying True Positive (TP) pixels.
The initial mask cannot deliver high enough confidence for those extended pixels.
Due to similar reasons, a combination of the two approaches is not beneficial, either.
\section{Results and Discussion}

Before discussing the results and illustrating the SILOP's efficacy, we first discuss the experimental setup used to generate the results discussed in this section.
The experiments are completed on a CentOS 7.9 Operating System executing on an Intel Core i7-8700 CPU with 16GB RAM and 2 Nvidia GeForce GTX 1080 Ti GPUs.
We executed our scripts with the following software versions: CUDA 11.5, Pytorch 3.7.4.3, torchvision 0.11.1, and Pytorch-lightning 1.5.1.
We use a ResNeSt101 DNN~\cite{RESNEST} as the backbone for PuzzleCAM to generate the unrefined predictions unless stated differently.
The PuzzleCAM module used in our framework is based on a GitHub clone of their original repository, followed by an extended hyper-parameter search, which identified the set parameters for better results.

For all experiments, the mean Intersection-over-Union (mIoU) ratio is used as the evaluation metric:

{
\begin{equation}
\centering
mIoU = \frac{1}{N} \sum_{i=1}^N \frac{ p_{i,i}}{\sum_{j=1}^N p_{i,j} + \sum_{j=1}^N p_{j,i} - p_{i,i}  }
\end{equation}
}

\noindent where $N$ is the total number of classes, $p_{i,i}$ the number of pixels classified as class $i$ when labelled as class $i$. $p_{i,j}$ and $p_{j,i}$ are the number of pixels classified as class $i$ that were labelled as class $j$ and vice-versa, respectively.

We used the PASCAL VOC2012 semantic segmentation benchmark for evaluating our framework.
It comprises 21 classes, including a background class, and most images include multiple objects.
Following the commonly adopted experimentation protocol for semantic segmentation, we use the $10,528$ augmented images, including the image-level labels, for training.
We use the validation set without augmentation to evaluate the network with $1,464$ examples.
Furthermore, our model is evaluated on the test set of $1,456$ images to ensure a constant comparison with the state-of-the-art. The test set does not contain any labels, and the evaluation results can only be obtained via the official VOC2012 evaluation website.

\subsection{Semantic Segmentation Performance on VOC2012}

\begin{table}[ht]
\caption{Performance in mIoU (\%) for PuzzleCAM and SILOP on the VOC2012 training dataset.\label{tab1}}

\centering 
{
\begin{tabular}{lccc}
\hline
Method       & CAM  & Aff. & Aff.+CRF \\ \hline
PuzzleCAM~\cite{PUZZLE}    & 59.9 & 69.5   &      69.7   \\
SILOP (Ours) & 61.1 & 70.0   & \textbf{70.2}        \\ \hline
\end{tabular}}
\end{table}

Table~\ref{tab1} compares the mIoU score of SILOP and the baseline, PuzzleCAM, when deployed with a ResNeSt101 backbone.
We observe that our framework achieves better prediction quality than the baseline, across all stages, even the AffinityNet stage.
Note that applying our PerimeterFit after AffinityNet does not lead to any improvements in the quality of the masks.
This implies that AffinityNet benefits from the higher-quality supervision offered by our framework, while PerimeterFit cannot improve if the borders are already drawn along some perimeters.
\begin{table}[ht]
\centering 
{
\caption{Comparison of SILOP with state-of-the-art techniques on the VOC2012 val and test datasets.\label{tab2}}
\begin{tabular}{lccc}
\hline
Method         & Backbone       & Val  & Test \\ \hline
Wang et al.~\cite{MCOF}          & ResNet-101     & 60.3 & 61.2 \\
Zeng et al.~\cite{ZENG}          & DenseNet-169   & 63.3 & 64.3 \\
Ahn et al.~\cite{IRNET}         & ResNet-50      & 63.5 & 64.8 \\ 
Jiang et al.~\cite{OOA}           & ResNet-101     & 63.9 & 65.9 \\
Fan et al.~\cite{ICD}           & ResNet-101     & 64.1 & 64.3 \\
Fan et al.~\cite{cian}          & ResNet-101     & 64.1 & 64.7 \\
Wang et al.~\cite{SEAM}          & WideResNet-38  & 64.5 & 65.7 \\
Lee et al.~\cite{FICKLE}        & ResNet-101     & 64.9 & 65.3 \\
Chang et al.~\cite{SUB}           & ResNet-101     & 66.1 & 65.9 \\
PuzzleCAM            & ResNeSt-101    & 66.5 & 67.3 \\
Lee et al.~\cite{AdvCAM}        & ResNet-50      & 68.1 & 68.0 \\
SILOP (Ours)        & ResNeSt-101    & 68.4 &  68.0 \\
Xie et al.~\cite{CLIMS}         & ResNet-50      & 70.4 & 70.0 \\ \hline
\end{tabular}}

\end{table}

\begin{table*}[ht]
\centering
\caption{Semantic segmentation performance on the first 11 classes of the VOC2012 training dataset for the final pseudo-labels. The bottom group contains results with CRF refinement, while the top group is without CRF. \label{tab3}}
{\begin{tabular}{lcccccccccccccccccccccc}

\hline
Method                              & bkg         & aero        & bike        & bird        & boat        & bottle      & bus         & car         & cat         & chair       & cow \\ \hline
AffinityNet~\cite{AFF}              & 88.2        & 68.2        & 30.6        & 81.1        & 49.6        & 61.0        & 77.8        & 66.1        & 75.1        & 29.0        & 66.0\\
Sub-Categories~\cite{SUB} (w/o CRF) & 88.1        & 49.6        & 30.0        & 79.8        & 51.9        &\textbf{74.6}&\textbf{85.1}& 73.7        & 85.1        &\textbf{31.0}& 77.6\\
PuzzleCAM (w/o CRF)                 & 88.5        &\textbf{78.5}& 43.4        &\textbf{88.6}& 61.4        &72.3         & 83.4        &\textbf{76.3}& 91.7        & 29.3        &\textbf{85.5}\\
SILOP (Ours) (w/o CRF)              &\textbf{89.0}& 78.3        &\textbf{44.4}& 88.1        &\textbf{66.0}& 73.8        & 84.0        & 73.8        &\textbf{92.2}& 30.6        & 83.1\\ \hline
MCOF~\cite{MCOF}                    & 87.0        & 78.4        & 29.4        & 68.0        & 44.0        & 67.3        & 80.3        & 74.1        & 82.2        & 21.1        & 70.7\\
Zeng et al.~\cite{ZENG}             &\textbf{90.0}& 77.4        & 37.5        & 80.7        & 61.6        & 67.9        & 81.8        & 69.0        & 83.7        & 13.6        & 79.4\\
FickleNet~\cite{FICKLE}             & 89.5        & 76.6        & 32.6        & 74.6        & 51.5        & 71.1        & 83.4        & 74.4        & 83.6        & 24.1        & 73.4\\
Sub-Categories~\cite{SUB} (w/ CRF)  & 88.8        & 51.6        & 30.3        & 82.9        & 53.0        &\textbf{75.8}&\textbf{88.6}& 74.8        & 86.6        &\textbf{32.4}& 79.9\\
PuzzleCAM (w/ CRF)                  & 88.6        &\textbf{79.2}& 43.7        &\textbf{89.0}& 61.8        & 72.1        & 83.3        &\textbf{76.0}& 92.0        & 29.5        & \textbf{86.0}\\
SILOP (Ours) (w/ CRF)               & 89.1        & 78.9        &\textbf{44.9}& 88.5        &\textbf{66.7}& 73.7        & 84.0        & 73.8        &\textbf{92.4}& 30.8        & 84.0
\end{tabular}}

\end{table*}

\begin{table*}[ht]
\centering

\caption{Semantic segmentation performance on the remaining 10 classes of the VOC2012 training dataset for the final pseudo-labels. The bottom group contains results with CRF refinement, while the top group is without CRF. \label{tab4}}
{\begin{tabular}{lcccccccccccccccccccccc}

\hline
Method                               & table          & dog           & horse       & motor        & person      & plant       & sheep       & sofa        & train       & tv          & mIoU\\ \hline
AffinityNet~\cite{AFF}               & 40.2           & 80.4          & 62.0        & 70.4         &\textbf{73.7}& 42.5        & 70.7        & 42.6        & 68.1        & 51.6        & 61.7\\
Sub-Categories~\cite{SUB} (w/o CRF)  & \textbf{53.2}  & 80.3          & 76.3        & 69.6         & 69.7        & 40.7        & 75.7        & 42.6        & 66.1        &\textbf{58.2}& 64.8\\
PuzzleCAM (w/o CRF)                  & 43.7           & \textbf{91.2} &\textbf{82.9}& \textbf{80.1}& 42.9        &\textbf{68.8}& 92.0        & 53.2        & 65.0        & 42.6        & 69.5\\
SILOP (Ours) (w/o CRF)               & 44.4           & 90.8          & 82.0        & 79.5         & 46.4        & 66.7        &\textbf{92.5}&\textbf{53.2}&\textbf{68.5}& 42.3        &\textbf{70.0}\\ \hline
MCOF~\cite{MCOF}                     & 28.2           & 73.2          & 71.5        & 67.2         & 53.0        & 47.7        & 74.5        & 32.4        & 71.0        & 45.8        & 60.3\\
Zeng et al.~\cite{ZENG}              & 23.3           & 78.0          & 75.3        & 71.4         & 68.1        & 35.2        & 78.2        & 32.5        &\textbf{75.5}& 48.0        & 63.3\\
FickleNet~\cite{FICKLE}              & 47.4           & 78.2          & 74.0        & 68.8         &\textbf{73.2}& 47.8        & 79.9        & 37.0        & 57.3        &\textbf{64.6}& 64.9\\
Sub-Categories~\cite{SUB} (w/ CRF)   & \textbf{53.8}  & 82.3          & 78.5        & 70.4         & 71.2        & 40.2        & 78.3        & 42.9        & 66.8        & 58.8        & 66.1\\
PuzzleCAM (w/ CRF)                   & 44.0           & \textbf{91.6} &\textbf{83.1}& \textbf{80.1}& 42.8        &\textbf{68.9}& 92.6        &\textbf{53.4}& 64.8        & 42.6        & 69.7\\
SILOP (Ours) (w/ CRF)                & 44.6           & 91.2          & 82.3        & 79.5         & 46.5        & 66.9        &\textbf{93.2}& 53.3        & 68.3        & 42.4        &\textbf{70.2}
\end{tabular}}

\end{table*}

Next, we compare our method with recent works using image-level supervision with or without a combination of some form of saliency in Table~\ref{tab2}.
Our framework uses PuzzleCAM as it was one of the most current approaches at the time.

SILOP performs better than most state-of-the-art approaches on both validation and testing sets.
Only the CLIMS approach reaches a higher mIoU score than our framework.
Note that since SILOP is the only approach utilizing edge detection, its use is orthogonal to state-of-the-art \cite{AdvCAM} or \cite{CLIMS}, and deploying them together may further increase the output quality.
Additionally, SILOP reaches a higher score than the more recent AdvCAM, despite using an outdated baseline.
Moreover, PuzzleCAM is the only approach that uses the powerful ResNeSt-101 backbone instead of ResNet-50 or ResNet-101, which makes its comparison to the state-of-the-art not fair.
Nevertheless, our approach is not dependent on the used backbone.

Next, we present a more comprehensive analysis and class-wise mIoU breakdown for all classes in the VOC2012 training dataset.
As seen in Table~\ref{tab2}, although SILOP achieves the highest quality predictions, on average, we observe in Table~\ref{tab3} and Table~\ref{tab4} that our framework might perform worse than the baseline for specific classes.
This hints that for some classes, SILOP removes a few true positives and reduces the class's performance.
SILOP loses around $2\%$ mIoU to PuzzleCAM on the classes \texttt{car} and \texttt{cow}.
We can explain the loss on the \texttt{car} class because it often overlaps with other surrounding objects and appears on images that tend to have many objects.
The inferiority on the \texttt{cow} class may be explained that PerimeterFit merges cows too often with grass or other objects.
In contrast, SILOP gains $4.9\%$, $3.7\%$, and $3.5\%$ over PuzzleCAM for the classes \texttt{boat}, \texttt{person}, and \texttt{train}, respectively.
The gain on the \texttt{boat} and the \texttt{train} classes can be explained by the fact that those are usually big objects that tend to be the only object in the image.
On the other side, the gain in the \texttt{person} class may stem from the fact that the overall quality of predictions of this class is relatively low compared to the other state-of-the-art methods, and there was much room for additional refinement.
SILOP is better than PuzzleCAM for $11$ out of $10$ classes.
However, on average, SILOP gains $2.2\%$  if it is better, whereas PuzzleCAN gains, on average, $0.9\%$ over SILOP when it is better.
Therefore, we conclude that if SILOP reduces the mask too much compared to the baseline, then we tend to lose not many true positives. At the same time, SILOP gains more mIoU points when it is successful in removing false positives and performs best when there is mostly a singular big object in the given image.

\subsection{Ablation studies}

\begin{table}[ht]0
\centering 
{
\caption{Comparison of $m_{FP}$ and $m_{FN}$ values of our framework and PuzzleCAM.\label{tab5}}
\begin{tabular}{lccc}
\hline
$m_{FP}$/$m_{FN}$       & CAM  & Aff. & Aff.+CRF \\ \hline
PuzzleCAM   & 0.35/0.35 & 0.28/0.18   & 0.27/\textbf{0.17}       \\
SILOP (Ours) & 0.27/0.39 & 0.26/0.19   & \textbf{0.25}/0.18        \\
\hline
\end{tabular}}

\end{table}

We will now analyze the prediction results of SILOP compared to the baseline, PuzzleCAM, at different stages of the framework to see where the improvements are derived.
We use the following metrics for evaluating the degree of over-/under-activation of a given prediction, based on previous state-of-the-art works~\cite{SEAM} to illustrate their efficacy:

{\fontsize{9pt}{9pt}\selectfont
\begin{equation}
m_{FP} = \frac{1}{C - 1}\sum^{C-1}_{y=1} \frac{FP_c}{TP_c},
m_{FN} = \frac{1}{C - 1}\sum^{C-1}_{y=1} \frac{FN_c}{TP_c}
\end{equation}}

\noindent where $TP_c$ denotes the number of true positive pixel predictions for each class $c$ of total $C$ classes. 
$FP_c$ and $FN_c$ denote the number of false positive and false negative predictions for their respective classes.
We exclude the background class from these metrics since it is an inverse of the foreground and nullify each other.
If the framework predicts a larger number of false positives, meaning that the framework highlights a larger area than the object, then the value of $m_{FP}$ will be higher.
On the other hand, if the framework predicts a larger number of false negatives, meaning that the framework highlights a smaller area than the object, then the value of $m_{FN}$ will be higher.

\begin{figure*}[t]
\centering
\includegraphics[width=150mm,scale=0.50]{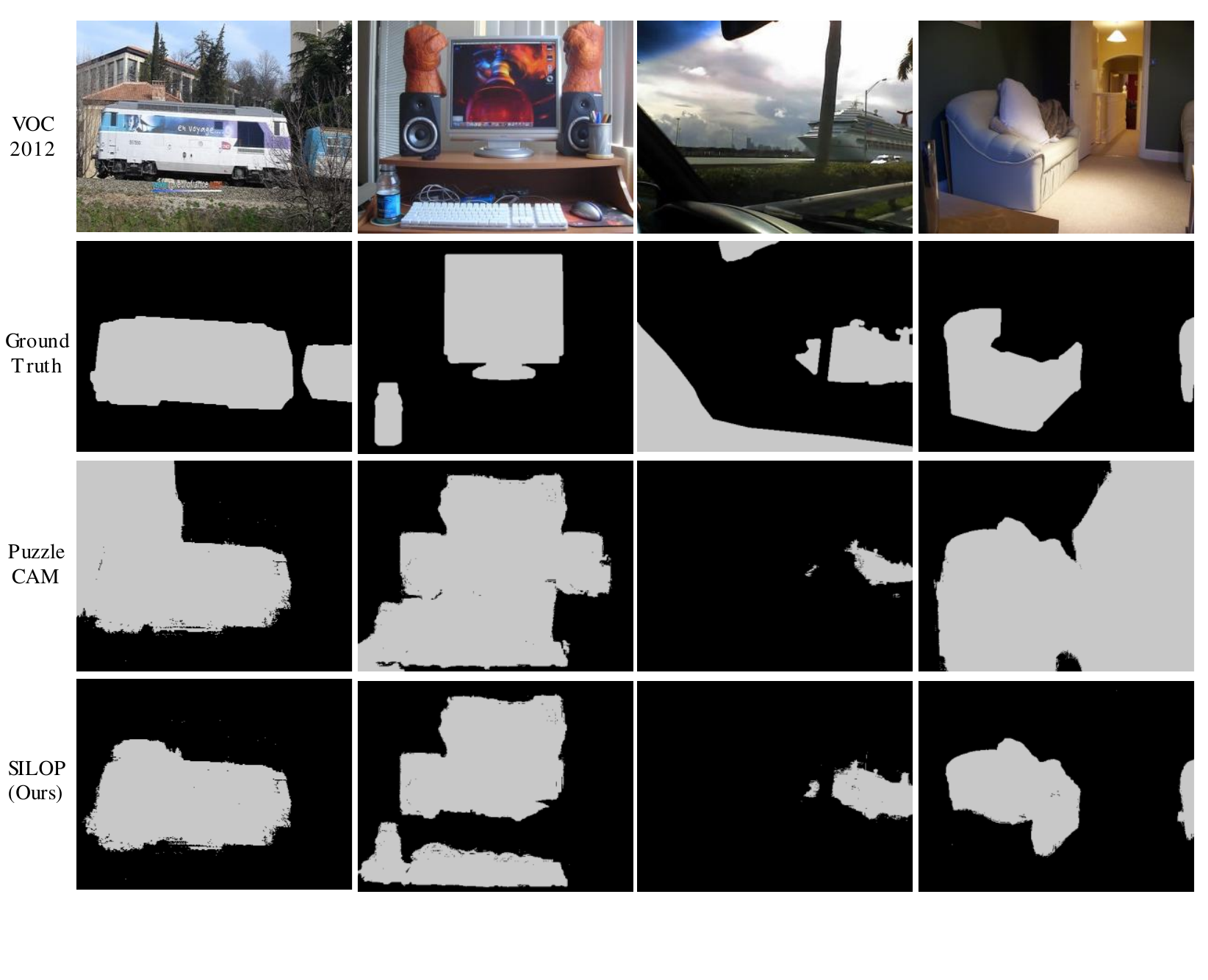} 
\caption{Example results of SILOP and PuzzleCAM for comparison. The first row shows the original VOC2012 images. The first image contains two objects of the class bus, the second a monitor and bottle, the third a ship and a car, and the fourth two couches. The second row shows the ground truths, the third row the results of the PuzzleCAM framework, and the last row the results of our framework.
\label{examples}}
\end{figure*}

When analyzing SILOP's and PuzzleCAM's predictions with these two metrics ($m_{FP}$ and $m_{FN}$), we have observed a reduction in the number of false-positive values but an increase in false-negative values for the proposed approach when compared to the baseline, as seen in Table~\ref{tab5}.
Furthermore, we notice that the improvement for $m_{FP}$ is larger than the decrease for $m_{FN}$, which was our framework's fundamental initial notion of improving prediction accuracy.
Based on our experimental results, AffinityNet achieves better results on a training dataset with low false positive rates than a dataset with overall higher quality but high FP rates.

Similarly, after training the AffinityNet model, we observe that the values of $m_{FP}$ still compare favorably to the baseline and that the value of $m_{FN}$ is slightly worse, as expected.
Two possible scenarios can explain this, (i)~the initial response map did not cover the complete object, in which case the framework finds the negatively predicted pixels inside the object and the Floodfill technique deletes this part of the object, or (ii)~the perimeter map does not detect the outlines of the object correctly, because of which parts of the object merge with the background, and the Floodfill method deletes it.
However, overall, the lower false positive rate outweighs the slightly higher false negative rate, as illustrated by the increase in prediction accuracy.
Overall, we observe that the approach's lower FP rate outweighs risking a slightly higher FN rate, as illustrated by the increase in prediction accuracy.

Fig.~\ref{examples} presents an overview of our experimental results.
The first row shows the original images from the VOC dataset, and the second row depicts their ground truths.
In the third row, we show the results of the baseline PuzzleCAM, and finally, the last row shows the results of SILOP.
We notice that for the first two images, our framework performs as intended in our motivation, namely that the prediction of the baseline PuzzleCAM still includes extra objects.
PerimeterFit removed the extra object in the first and last image and some in the second.
That is the expected behavior that our framework cannot find the missing objects in the first and third images.
Surprisingly, the mask of SILOP covers more of the target object in the third image than PuzzleCAM.
We could explain this behavior using a higher threshold for the rough CAM.
We included more of the object, and the PerimeterFit module deleted fewer parts of the image than AffinityNet when provided with worse pseudo-labels.

\section{Conclusion}

In this paper, we have proposed our SILOP framework that introduces the novel PerimeterFit module between CAM and the Affinitynet model to address the weaknesses of pixel-similarity-based refinement methods.
The PerimeterFit module provides additional saliency by utilizing unsupervised semantic segmentation models and edge detection methods, which refine CAM predictions to obtain higher-quality training labels for state-of-the-art FSSS models.
The PerimeterFit module can be incorporated into any pre-existing image-level-based semantic segmentation framework to boost the quality of its predictions.
We showed that the predictions generated by SILOP achieve state-of-the-art performance on the VOC2012 dataset, which proves its effectiveness compared to other approaches.
Our framework is open-source to ensure reproducible research and accessibility.
The source code is online at \url{https://github.com/ErikOstrowski/SILOP}.

\section*{Acknowledgments}
This work is part of the Moore4Medical project funded by the ECSEL Joint Undertaking under grant number H2020-ECSEL-2019-IA-876190.
This work was also supported in parts by the NYUAD’s Research Enhancement Fund (REF) Award on “eDLAuto: An Automated Framework for Energy-Efficient Embedded Deep Learning in Autonomous Systems”, and by the NYUAD Center for Artificial Intelligence and Robotics (CAIR), funded by Tamkeen under the NYUAD Research Institute Award CG010.

\bibliography{our_framework}
\bibliographystyle{ieeetr}

\vspace{12pt}
\color{red}

\end{document}